\documentclass{article} 
\usepackage{iclr2027_conference,times}

\usepackage{amsmath,amsfonts,bm}

\def\eqref#1{equation~\ref{#1}}

\def\1{\bm{1}}

\DeclareMathAlphabet{\mathsfit}{\encodingdefault}{\sfdefault}{m}{sl}
\SetMathAlphabet{\mathsfit}{bold}{\encodingdefault}{\sfdefault}{bx}{n}

\usepackage{enumitem}
\usepackage{url}
\usepackage{hyperref}
\usepackage{url}
\usepackage{booktabs}
\usepackage{array}
\usepackage{multirow}
\usepackage{graphicx}
\usepackage{tikz}
\usepackage{pgfplots}
\usepgfplotslibrary{groupplots}
\pgfplotsset{compat=1.18}
\usepackage{xcolor}
\usepackage{listings}
\usepackage{fontawesome5} 
\usepackage{twemojis}     

\DeclareTextSymbolDefault{\textquotedbl}{T1}
\definecolor{lstback}{gray}{0.97}
\definecolor{lstrule}{gray}{0.75}
\title{SpecGuardBench: Can Coding Agents Construct Reliable Validation Harnesses from Natural Language Specifications?}

\author{%
Zhiqing Zhong$^{1}$, Mingyu Chen$^{1}$, Aoyang Fang$^{1}$, Yuzhong Zhang$^{1}$, \\
\textbf{Haoye Tian$^{2}$, Lionel Briand$^{3,4,5}$, Boxi Yu$^{3,4}$, Pinjia He$^{1}$\thanks{Pinjia He is the corresponding author.}} \\
$^{1}$The Chinese University of Hong Kong, Shenzhen \quad $^{2}$Aalto University \\ $^{3}$Lero, the Research Ireland Centre for Software \quad $^{4}$University of Limerick \\ $^{5}$University of Ottawa \\
\texttt{\{zhiqingzhong,mingyuchen1,aoyangfang\}@link.cuhk.edu.cn} \\
\texttt{yuzhongzhang@link.cuhk.edu.cn}, \texttt{haoye.tian@aalto.fi} \\ \texttt{\{lionel.briand,boxi.yu\}@lero.ie}, \texttt{hepinjia@cuhk.edu.cn} \\ [0.6em]
\hypersetup{pdfborder={0 0 0}}
\faGithub~\href{https://github.com/coxhin/SpecGuardBench}{\nolinkurl{coxhin/SpecGuardBench}} \qquad
\raisebox{-0.2em}{\twemoji[height=1.15em]{1f917}}~\href{https://huggingface.co/datasets/coxhin/SpecGuardBench}{\nolinkurl{coxhin/SpecGuardBench}}
}

\newcommand{\benchname}{\textsc{SpecGuardBench}}
\newcommand{\refto}{\,\textcolor{black!45}{$\rightarrow$}\,}
\newcommand{\zp}{\phantom{0}}

\iclrfinalcopy 
\begin{document}

\maketitle
\ificlrfinal\lhead{Preprint}\fi

\begin{abstract}
As coding agents become more capable, tests remain the primary evidence for judging whether their code is correct. Yet passing available tests does not establish that an implementation satisfies its natural language specification. Reliable validation requires checks that faithfully capture the specification and scenarios that thoroughly exercise the code, but current benchmarks provide limited insight into whether agents can construct such evidence themselves.
We introduce \benchname{}, a benchmark of 291 tasks from 13 real-world repositories, covering 8 types of verification logic such as functional postconditions and interaction protocols. Given a natural language specification, repository context, and a reference implementation, an agent constructs a validation harness comprising executable checks and test scenarios. The checks should reject implementations that violate the specification while accepting those that satisfy it, and the scenarios should thoroughly exercise the code.
Across four proprietary and open-weight coding agents, none constructs a harness scored as faithful on the benchmark's implementations for more than 63\% of the tasks. The bottleneck lies less in exploring test scenarios than in constructing checks that faithfully reflect the specification. Specifically, the agents show a clear tendency to write checks that mirror the reference implementation rather than the specification, and therefore reject implementations that the specification permits.

\end{abstract}
\section{Introduction}
As coding agents take on increasingly complex changes and long-horizon tasks across software repositories, natural language specifications from humans or agents guide what they should accomplish and provide a basis for judging their outputs. Reliable validation of whether the code aligns with specifications is therefore essential and challenging for benchmark evaluation, interactive coding workflows, and reinforcement learning with verifiable rewards (RLVR). Across these scenarios, finite executable test suites remain a primary mechanism for validation and supervision~\citep{jimenez2024swe,ma2026rethinking}. Yet passing a test suite does not establish conformance to the underlying specifications: an incorrect implementation may satisfy the tested examples while violating specifications in untested situations~\citep{liu2023your,10.1145/3803437.3805574}.
Moreover, scaling test suites is challenging, as test inputs are typically coupled with expected outcomes.
A finite test suite is thus bound to be incomplete, making it an unreliable validation paradigm.

Reliable validation requires checks that \textbf{faithfully capture} the stated specifications and the exploration of test scenarios that \textbf{thoroughly exercise} the relevant code~\citep{6963470}. 
Faithful checks should identify specification violations without rejecting behavior that the specifications permit, while thorough exploration of test scenarios should apply these checks to executions spanning a broad range of paths and program states.
Although exhaustive testing is generally infeasible, exploration budget can be scaled to application-specific reliability requirements, with additional compute helping reduce the risk of undetected faults~\citep{10.1145/3368089.3409729}.
Expressing specifications as reusable executable checks decoupled from test scenarios lets the same checks assess newly generated test scenarios and real-world executions, supporting continued validation beyond the initial test suite and after deployment.

Existing benchmarks increasingly assess models’ ability to construct evidence for code validation, but only partially address whether the resulting evidence supports scalable, faithful validation against natural language specifications.
Previous test-generation evaluations typically ask whether LLMs can generate tests that improve fault detection~\citep{jain2025testgeneval,jing2026pbt}, reproduce bugs~\citep{3737916.3740517,ahmed2024tdd}, or exercise more program structure~\citep{wang-etal-2025-testeval}. However, these evaluations overlook the risk that generated tests overfit the reference implementation rather than faithfully capture the stated specifications.
Verification-oriented benchmarks~\citep{ye2025verina,thakur2026clever} move beyond this paradigm by asking agents to construct machine-checkable specifications that characterize intended program behavior. 
Machine-checkable specifications are typically decoupled from specific inputs and exercised by a formal verifier.
However, they often evaluate these specifications within predefined formal languages and verification interfaces, providing limited evidence about agents’ ability to validate code against diverse natural language specifications in real-world repositories. 
Such scenarios may involve complex dependencies and various types of verification logic (e.g., functional postconditions and interaction protocols).

\begin{figure}[t]
    \centering
    \includegraphics[width=1\linewidth]{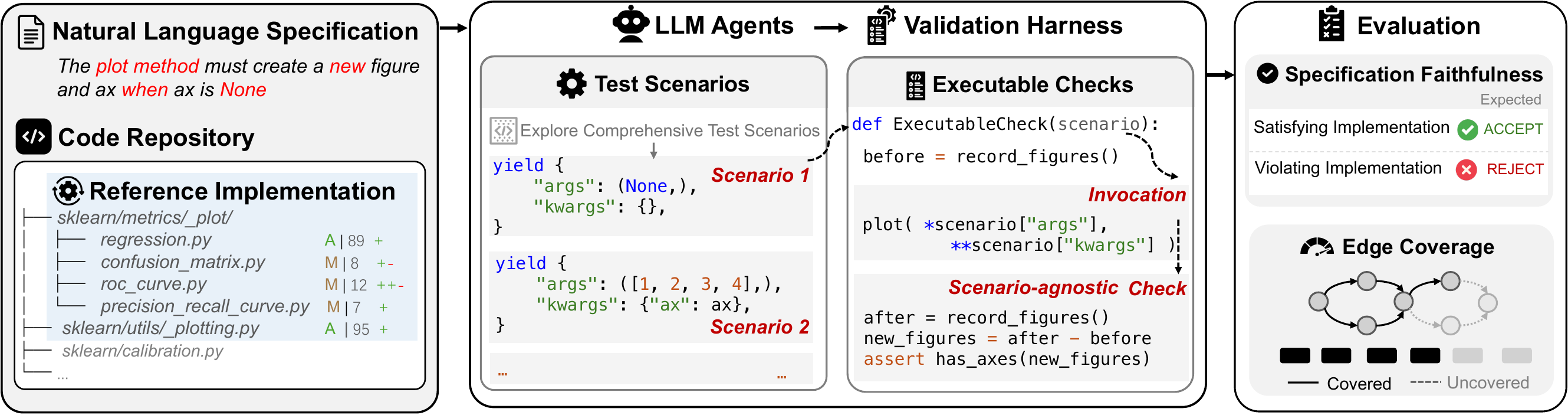}
    \caption{Overview of \benchname{} task. Given a natural language specification, the repository, and a reference implementation, an agent constructs a validation harness of test scenarios and executable checks; the harness is judged on whether it accepts implementations that satisfy the specification, rejects those that violate it, and how thoroughly the scenarios exercise the relevant code.}
    \label{fig:task_formulation}
\end{figure}

To bridge this gap, we introduce \benchname{}, which evaluates whether LLM agents can construct evidence that supports thorough, faithful validation against natural language specifications in real-world repositories. 
\benchname{} consists of 291 validation problems, covering eight types of verification logic. 
As shown in Figure~\ref{fig:task_formulation}, given a natural language specification, repository context, and a reference implementation, an agent constructs a validation harness comprising comprehensive test scenarios and executable checks. 
The test scenarios supply concrete arguments that drive the code under validation through a broad range of program states. The executable checks perform invocations of the code under validation with whatever arguments they receive and assert the properties that the natural language specification requires of the resulting behavior, so the same checks apply across executions, even after deployment.
Together, the checks and the scenarios should thoroughly exercise specification-relevant code and expose violations without rejecting implementations that satisfy the specification.

Using \benchname{}, we evaluate four coding agents spanning proprietary and open-weight models, and none constructs a validation harness scored as faithful on the benchmark's implementations for more than 63\% of the tasks.
The primary bottleneck is the faithfulness of the checks, not the thoroughness of the scenarios.
The agents detect the violations on most tasks, but their checks also assert requirements the specification does not impose, and the resulting false alarms leave every agent's faithfulness 11 to 22 percentage points below its detection rate.
Even when given scenarios that expose the violation, the checks of three of the four agents still accept most of the violations they missed, and the scenarios account for at most 33 to 42\% of each agent's failures.
Moreover, the coding agents ground their checks in the reference implementation rather than in the specification: when shown a violating reference, every agent loses 19 to 31 points of faithfulness, because its validation harness then accepts the violation or rejects the gold implementation.
These results reveal a gap in coding agents' ability to validate code against a natural language specification.
Closing this gap depends above all on checks that assert no more and no less than the specification requires, which none of the agents we evaluate reliably produces.

In summary, we make the following contributions:
\begin{itemize}[topsep=5pt]
\item We introduce \benchname{}, a benchmark that evaluates whether LLM agents can construct reliable evidence to validate code in real-world repositories against natural language specifications.
\item We release a scalable data construction pipeline that converts code with relevant documentation into additional evaluation tasks.
\item We evaluate four proprietary and open-weight LLMs and analyze the remaining challenges in depth.
\end{itemize}

\section{Related Work}

\paragraph{Test generation.}
Prior evaluations of LLM test generation score generated tests by the faults they detect, the reported bugs they reproduce, or the coverage they reach, and when a test is validated, it is validated only against the one correct implementation available, while overlooking whether the test overfits that implementation. TestEval~\citep{wang-etal-2025-testeval} and TestGenEval~\citep{jain2025testgeneval} ask a model to write concrete test cases for a given program; the former measures the line, branch, and path coverage the tests reach, and the latter additionally measures the fraction of injected faults they detect, while SWT-Bench~\citep{3737916.3740517} and TDD-Bench Verified~\citep{ahmed2024tdd} give a reported issue together with its repository and score an instance as solved when some generated test fails before the fix and passes after it. Since each concrete test couples one input with its expected outcome, it offers no reusable check for other inputs, and thoroughness can only be measured by the number of tests or the coverage they reach. Property-based testing decouples the two by stating a property once and generating inputs separately, and prior studies ask whether LLMs can write such properties from a library's documentation and code~\citep{vikram2024largelanguagemodelswrite,maaz2025agentic}. PBT-Bench~\citep{jing2026pbt} is the closest to our task in form. Given a Python library with injected bugs and its documentation, an agent writes Hypothesis tests, each pairing a property drawn from the documentation with a random input distribution, and a bug counts as found when some test fails on the buggy library and passes on the fixed one. This design leaves two limitations. First, the fixed library is the only implementation a test must accept, so a test that overfits it still counts as finding a bug. Second, the agent describes only an input space, from which the testing framework draws random samples under a fixed budget, and is told to leave finding the violation to the sampler. The task thus limits how far the agent explores and constructs the scenarios that reach a violation, and whether a bug is found depends in part on random sampling, a process the agent neither takes part in nor observes. \benchname{} fixes one atomic specification per task, scores the harness as a whole by its verdict on each implementation, and includes valid alternative implementations that satisfy the specification but differ from the reference in code the specification does not constrain, so that checks fitted to the reference rather than to the specification are exposed, and asks the agent to explore and construct the scenarios itself.

\paragraph{Specification generation.}
A second line of work asks a model to turn a natural language description into a machine-checkable specification in a fixed formalism, typically a pre- and postcondition contract, and scores the specification by whether it accepts correct behavior and rejects incorrect behavior. In formal settings, the specification is written in a verification language and checked by a verifier. Verina~\citep{ye2025verina} and CLEVER~\citep{thakur2026clever} generate Lean specifications from a natural language description and compare them with a ground-truth specification by proving implication or equivalence between the two, while nl2spec~\citep{10.1007/978-3-031-37703-7_18} targets temporal logic. In executable settings, the specification is a set of runtime assertions, checked by execution rather than by formal proof. nl2postcond~\citep{10.1145/3660791} generates postconditions, assertions over a function's inputs and return value, from its docstring, and scores them by whether the reference implementation satisfies them on the test inputs and by the fraction of incorrect programs they reject. CodeSpecBench~\citep{chen2026codespecbenchbenchmarkingllmsexecutable} extends this to preconditions, assertions over the inputs, and to repository-level tasks drawn from SWE-bench. It scores a specification by whether it accepts executions of a reference implementation and rejects illegal inputs and outputs that differ from the reference implementation. A related line infers specifications from the implementation itself, but since the implementation is then the standard, a specification that mirrors it is the intended outcome rather than a failure~\citep{10.1109/ICSE55347.2025.00129,DBLP:conf/cav/WenCSXQHLCT24,le-cong-etal-2025-llms}. These benchmarks differ from ours in three respects. The formal line requires both the code and the specification to be written in a verification language, with ground truth formalized by hand, which limits it to standalone functions and excludes real repositories with their dependencies. The executable line reaches real repositories, but the inputs on which a specification is exercised are a curated test set supplied by the benchmark, so the model is never asked to explore the code and its thoroughness is never measured. Above all, most of these benchmarks define correct behavior by one implementation or by tests derived from it, and thus also overlook the risk that a specification overfits the reference implementation. \benchname{} therefore asks the agent for both the checks and the scenarios, measures how thoroughly the scenarios exercise the relevant code, and scores the checks against valid alternative implementations, which distinguishes checks faithful to the specification from checks faithful to the reference implementation.

\section{\benchname{}}


\benchname{} measures whether an agent can turn a natural language specification into evidence that supports thorough, faithful validation. Its tasks are produced by an automated pipeline built on top of FeatureBench~\citep{zhou2026featurebench}, described in Section~\ref{sec:collection}.

\subsection{Task Formulation}
\label{sec:task}

Each task provides the agent with one atomic natural language specification, a reference implementation containing the specified functionality, and the code repository. The agent is not told whether the reference implementation aligns with the specification. It is therefore instructed to take the specification, not the implementation, as the standard and to produce executable checks that reject every violating implementation and accept every satisfying one, together with test scenarios that exercise the relevant code as thoroughly as possible. The agent submits a validation harness comprising two components: a generator of test scenarios, each specifying the concrete arguments of a single invocation of the code under validation, and scenario-agnostic executable checks that, given one scenario, perform the invocation and assert the properties required by the specification. Appendix~\ref{app:task} gives the contract that fixes the form of the two components, and a complete task.

We then evaluate the validation harness by running its checks against three kinds of implementations: the gold implementation, which satisfies the specification by construction; a violating implementation constructed for that specification; and valid alternative implementations, which modify only code that the specification does not constrain. Section~\ref{sec:collection} describes the construction of the latter two. We measure the faithfulness rate, the fraction of tasks whose checks reject the violating implementation and accept every satisfying one, and report its two conditions separately. The detection rate is the fraction of tasks whose checks reject the violating implementation, and the false alarm rate is the fraction whose checks reject the gold implementation or a valid alternative. Since no evaluation can exhaust the implementations that satisfy or violate a specification, the faithfulness rate measured on these implementations is an empirical upper bound. This bound errs only in the agents' favor. A validation harness faithful to the specification is always scored as faithful, while every failure we report is witnessed by a concrete implementation that the harness misjudges, so our results can only understate how far the agents fall short. We additionally record the edge coverage attained by the generated scenarios, which indicates how thoroughly they explore the relevant code and, when the checks pass, helps assess the residual risk of unexercised violation~\citep{10.1007/3-540-45732-1_17,1044339}. Edge coverage is the fraction of control-flow edges the scenarios execute in the gold-patch functions that implement the specification and in every gold-patch function reachable from them by calls.

\subsection{Benchmark Construction}
\label{sec:collection}

\subsubsection{Source Instances}
Each validation task needs a natural language specification, code known to align with it, and code known to violate it. We build \benchname{} on FeatureBench, which already provides high-quality natural language specifications that align with the code implementing them. On top of each FeatureBench instance, we construct the violating implementations, as Figure~\ref{fig:overview_specbench} illustrates and the rest of this section describes. Each FeatureBench instance targets a feature of an open-source Python project and ships its specification as a task statement and interface descriptions describing the feature, together with the gold patch that implements it and the test suite that validates it. Compared with issue reports in bug-fixing benchmarks such as SWE-bench~\citep{jimenez2024swe}, which describe a bug symptom and leave the intended behavior implicit, a FeatureBench specification explicitly states a feature's intent, which executable checks must validate. From FeatureBench, we randomly sample 30 instances that do not require a specific GPU environment, covering 13 repositories.

\subsubsection{Specification Decomposition}
A FeatureBench specification describes a whole feature and therefore includes many requirements, each of which an implementation can satisfy or break on its own, so each is a separate place to go wrong and a separate point to verify. We use an LLM (Claude Opus 5) to split it until each piece is a requirement that can be violated without violating another, and call these pieces atomic specifications (Figure~\ref{fig:overview_specbench}a). Each atomic specification must be copied from the original specification rather than inferred from the gold patch, is labeled as behavioral, structural, or an implementation constraint, and is paired with its implementation in the gold patch.

\subsubsection{Violation Synthesis}
\begin{figure}
    \centering
    \includegraphics[width=1\linewidth]{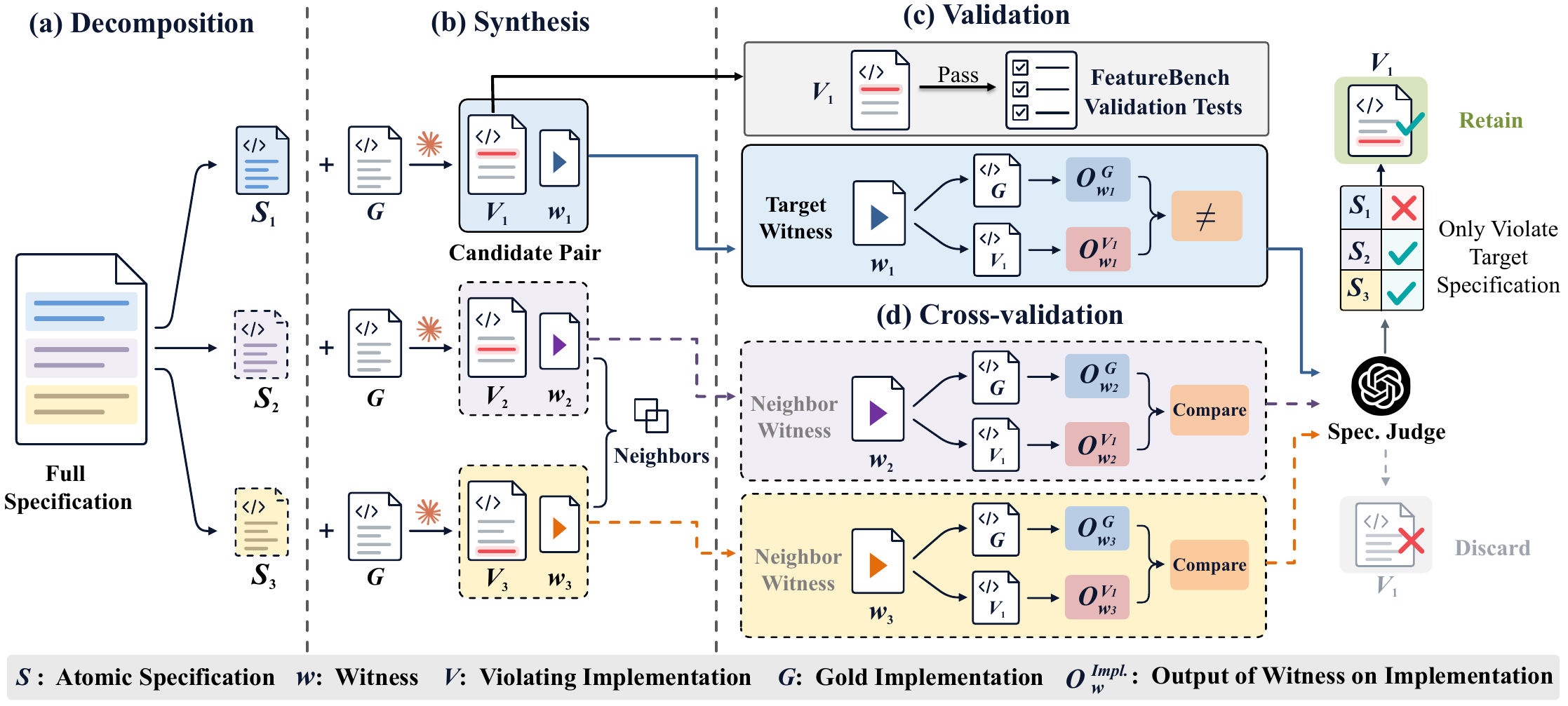}
    \caption{Construction of a \benchname{} task. An LLM decomposes the FeatureBench specification into atomic specifications (a) and, for each, writes a candidate violating implementation with a differential witness (b). The candidate must pass the instance's original test suite and diverge from the gold implementation on its witness (c), and the witnesses of neighboring specifications are run against it (d). A candidate is kept only when a second LLM confirms from these divergences that it violates the targeted specification alone.}
    \label{fig:overview_specbench}
\end{figure}

For each atomic specification, we then build the violating implementation that the checks must catch (Figure~\ref{fig:overview_specbench}b). Given the targeted atomic specification and the other specifications of the FeatureBench instance, the same LLM produces a candidate by injecting a violation of the targeted specification into its implementation, so that this atomic specification no longer holds while the others still do. The violation should produce an effect observable by executing the code rather than inspecting its source, so the task is solvable by executable checks, and the divergence should be deterministic rather than an effect of environmental noise. As evidence of the violation, the LLM also writes a differential witness: scenarios whose execution produces such an effect and diverges from the gold implementation.

The pipeline discards a candidate that fails either of two filters (Figure~\ref{fig:overview_specbench}c). First, we run it against the FeatureBench instance's validating test suite. A surviving candidate is one the original test suite cannot catch, since it violates the specification only in situations the suite never exercises. It is therefore challenging by construction, and lies precisely in the gap between passing tests and conforming to the specification that motivates this work. Second, we run the witness against the gold implementation and against the candidate, and keep the candidate only when both runs finish and their outputs differ. A second LLM from a different model family (GPT-5.6-sol) then checks the divergence and confirms, by majority vote over three independent samples, that it indicates a genuine violation of the specification rather than an incidental difference.
We use two LLMs from different model families, one to synthesize a candidate and the other to check it, to mitigate the ambiguity of the natural language specification: a candidate is kept only when both agree that it violates the target specification without violating the others.

\subsubsection{Keeping the Other Specifications Satisfied}
We then check that the candidate violates only the targeted atomic specification, so that the other specifications of the FeatureBench instance still hold (Figure~\ref{fig:overview_specbench}d). Such a candidate still satisfies every other specification of the instance, so it can later test whether the checks written for those specifications overfit and reject an implementation they should accept. We check this in two complementary ways. We call a specification's neighbors the other specifications whose implementation overlaps with its implementation, and run each of their witnesses against the candidate under check to see whether it diverges from the gold implementation, suggesting the neighboring specification no longer holds. Given the candidate and any divergence, the second LLM then decides for every other specification of the instance whether it still holds, and rejects the candidate by the same majority vote if any of them no longer holds. We keep an atomic specification only when both its own candidate and that of at least one neighboring specification survive this check, the neighbor's candidate serving as a valid alternative implementation that faithful checks should not reject.

To validate these two LLM judgments, we manually reviewed a sample of 50 tasks from the final benchmark, stratified by instance. Two authors independently judged, for each task, whether the violating implementation violates the targeted atomic specification and whether each valid alternative implementation still satisfies it. The two annotators agreed on 94.7\% of judgments, and their adjudicated judgment confirmed the pipeline's decision on 98\% of tasks.

\input{figures/bench_statistics}

\subsection{Benchmark Statistics}
\label{sec:composition}

Across the 30 instances, our decomposition and curation pipeline yields 291 atomic specifications, each with at least one valid alternative implementation. These tasks constitute the \textbf{Full Set} of \benchname{}.
As shown in Figure~\ref{fig:instance-count}, the task counts vary considerably across instances, with a median of 7.5 and an interquartile range from 4 to 11 tasks. 
To facilitate wider adoption of \benchname{}, we construct a \textbf{Lite Set} of 100 specifications by random sampling stratified by instance: every instance contributes at least one specification, and the remaining slots are allocated in proportion to each instance's share of the full set.

We further performed open coding~\citep{6092576} on a stratified sample of the specifications and their differential witnesses to examine whether the underlying properties map to existing formal verification logics. Through this process, we derived eight types of verification logic, ranging from postconditions on a return value to frame conditions, interaction protocols, and relations across executions, and each corresponds to a construct of an existing formal verification logic.
Figure~\ref{fig:spec-category} shows their distribution across \benchname{}. Details are provided in Appendix~\ref{app:logic}.
\section{Experiments and Analysis}
\label{sec:experiments}

On \benchname{}, none of the four coding agents we evaluate produces faithful validation harnesses for more than 63\% of the 291 tasks, and detection exceeds faithfulness by 11--22 points for every agent. Section~\ref{sec:setup} describes the setup and Section~\ref{sec:main-results} reports the main results. Section~\ref{sec:failure} locates the failures. False alarms come from checks that assert more than the specification requires, and most missed violations persist even when we replace the scenarios with ones proven to expose the violation. Section~\ref{sec:ablations} varies two conditions of the main experiment: which implementation is shown as the reference, and how many model calls the agent may make.

\subsection{Setup}
\label{sec:setup}

We evaluate four coding agents spanning proprietary and open-weight models: Codex with GPT-5.6-sol, Claude Code with Claude Opus 5, OpenCode with DeepSeek-V4-Pro, and OpenCode with GLM-5.3. Each agent runs with the default settings of its agent harness under a step budget of 80 model calls per task. The reference implementation shown to the agent is the gold implementation.

\subsection{Main Results}
\label{sec:main-results}

\begin{table}[t]
\centering
\caption{Performance of four baseline coding agents on the 291 validation tasks of \benchname{}. Cost denotes the average cost per task in U.S. dollars.}
\label{tab:main}
\small
\begin{tabular}{lccccc}
\toprule
Agent & Detection $\uparrow$ & False Alarm $\downarrow$ & Faithfulness $\uparrow$ & Edge Cov. $\uparrow$ & Cost \\
\midrule
Codex / GPT-5.6-sol         & \textbf{81.8\%} & 23.0\% & 61.9\% & 82.0\% & \$0.74 \\
Claude Code / Claude Opus 5 & 65.3\% & 20.6\% & 50.2\% & 72.4\% & \$1.79 \\
OpenCode / DeepSeek-V4-Pro  & 66.7\% & 26.8\% & 45.0\% & 75.1\% & \$0.43 \\
OpenCode / GLM-5.3          & 74.2\% & \textbf{15.1\%} & \textbf{62.9\%} & \textbf{84.3\%} & \$1.48 \\
\bottomrule
\end{tabular}
\end{table}

Table~\ref{tab:main} reports the performance of the four coding agents we evaluate, and shows that the task is challenging not only in detecting implementations that violate the specification, but also in not rejecting those that satisfy it. Faithfulness, which requires both, ranges from 45\% to 63\% of the 291 tasks, so even the two most faithful agents fall short on more than a third of them. The agents miss the violation on 18\% to 35\% of the tasks and, sometimes on the same tasks, reject an implementation that satisfies the specification on 15\% to 27\%. Detection is the quantity a test-generation evaluation would report, and such an evaluation would underestimate the false alarm rate and rank the agents by differences that need not carry over to faithfulness. GPT-5.6-sol detects the violation on eight points more tasks than GLM-5.3 but also raises false alarms on eight points more, so the two agents reach a similar level of faithfulness (Appendix~\ref{app:feature-level} reports the scores averaged over features). Conversely, DeepSeek-V4-Pro detects as often as Claude Opus 5 but raises false alarms on six points more tasks. The two most faithful agents are also the two with the highest edge coverage, consistent with checks exercised through more thorough scenario exploration.

\subsection{Failure Analysis}
\label{sec:failure}

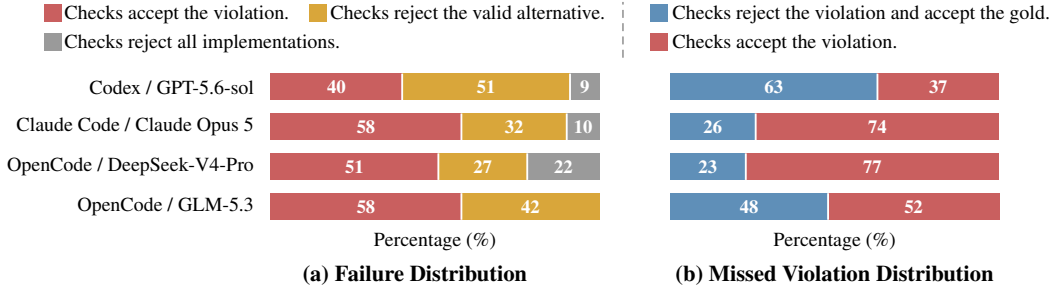
\begin{figure}[t]
\centering
\begingroup
\definecolor{missScenario}{HTML}{5F8FB8}
\definecolor{missChecks}{HTML}{C95F5F}
\definecolor{missAlternative}{HTML}{D9A63A}
\definecolor{missRejectAll}{HTML}{9A9A9A}
\definecolor{missText}{HTML}{000000}
\newcommand{\missAgentFont}{\rmfamily\fontsize{7.5}{8.5}\selectfont}
\newcommand{\missAxisFont}{\rmfamily\fontsize{7.5}{8.5}\selectfont}
\newcommand{\missTitleFont}{\rmfamily\fontsize{8.5}{8.5}\selectfont}
\begin{tikzpicture}[
    x=0.95\linewidth,
    y=1pt,
    font=\rmfamily\fontsize{7}{8.5}\selectfont,
    text=missText,
    legend/.style={anchor=north west, align=left, inner sep=0pt,
        font=\rmfamily\fontsize{7.5}{7.2}\selectfont},
    agent label/.style={anchor=east, inner sep=0pt,
        font=\missAgentFont},
    axis label/.style={inner sep=0pt, text=black,
        font=\missAxisFont},
    percentage/.style={font=\rmfamily\bfseries\fontsize{7}{8}\selectfont, inner sep=0pt},
    panel title/.style={anchor=north, align=center, inner sep=0pt,
        font=\missTitleFont}
]
\path[use as bounding box] (0,-14) rectangle (1,95);
\begin{scope}[xshift=-0.02\linewidth]

\newcommand{\missbar}[4]{%
    \pgfmathsetmacro{\firstEnd}{#1}
    \pgfmathsetmacro{\secondEnd}{#1+#2}
    \pgfmathsetmacro{\thirdEnd}{#1+#2+#3}
    \begin{scope}[xshift=-0.025\linewidth]
    \ifnum#1>0\relax
        \fill[missChecks]
            (0.300,#4-5)
            rectangle ({0.300+0.00328*\firstEnd},#4+5);
        \node[percentage, text=white]
            at ({0.300+0.00328*#1/2},#4) {#1};
    \fi
    \ifnum#2>0\relax
        \fill[missAlternative]
            ({0.300+0.00328*\firstEnd},#4-5)
            rectangle ({0.300+0.00328*\secondEnd},#4+5);
        \node[percentage, text=white]
            at ({0.300+0.00328*(\firstEnd+#2/2)},#4) {#2};
        \draw[white, line width=0.7pt]
            ({0.300+0.00328*\firstEnd},#4-5)
            -- ({0.300+0.00328*\firstEnd},#4+5);
    \fi
    \ifnum#3>0\relax
        \fill[missRejectAll]
            ({0.300+0.00328*\secondEnd},#4-5)
            rectangle ({0.300+0.00328*\thirdEnd},#4+5);
        \node[percentage, text=white]
            at ({0.300+0.00328*(\secondEnd+#3/2)},#4) {#3};
        \draw[white, line width=0.7pt]
            ({0.300+0.00328*\secondEnd},#4-5)
            -- ({0.300+0.00328*\secondEnd},#4+5);
    \fi
    \end{scope}
}
\begin{scope}[yshift=-5pt]
\draw[
    black!50,
    densely dashed,
    line width=0.5pt
] (0.625,75) -- (0.625,96);

\fill[missScenario] (0.652,88.5) rectangle (0.669,95);
\node[legend] at (0.674,95)
    {Checks reject the violation and accept the gold.};

\fill[missChecks] (0.652,77.5) rectangle (0.669,84);
\node[legend] at (0.674,84)
    {Checks accept the violation.};

\begin{scope}[xshift=-0.04\linewidth]
\fill[missChecks] (0.090,88.5) rectangle (0.107,95);
\node[legend] at (0.110,95) {Checks accept the violation.};
\fill[missAlternative] (0.355,88.5) rectangle (0.372,95);
\node[legend] at (0.375,95) {Checks reject the valid alternative.};
\fill[missRejectAll] (0.090,77.5) rectangle (0.107,84);
\node[legend] at (0.110,84) {Checks reject all implementations.};
\end{scope}
\end{scope}
\foreach \agent/\rowY/\missedPct/\alternativePct/\rejectAllPct/\repairedPct in {
    {Codex / GPT-5.6-sol}/59/40/51/9/63,
{Claude Code / Claude Opus 5}/44/58/32/10/26,
{OpenCode / DeepSeek-V4-Pro}/29/51/27/22/23,
{OpenCode / GLM-5.3}/14/58/42/0/48
} {
    \node[agent label, xshift=-0.025\linewidth] at (0.284,\rowY) {\agent};
    \missbar{\missedPct}{\alternativePct}{\rejectAllPct}{\rowY}
    \pgfmathtruncatemacro{\repairedPctInt}{\repairedPct}
    \pgfmathtruncatemacro{\persistentPct}{100-\repairedPct}
    \fill[missScenario] (0.672,\rowY-5)
        rectangle ({0.672+0.00328*\repairedPctInt},\rowY+5);
    \fill[missChecks] ({0.672+0.00328*\repairedPctInt},\rowY-5)
        rectangle (1,\rowY+5);
    \draw[white, line width=0.7pt]
        ({0.672+0.00328*\repairedPctInt},\rowY-5)
        -- ({0.672+0.00328*\repairedPctInt},\rowY+5);
    \node[percentage, text=white]
        at ({0.672+0.00328*\repairedPctInt/2},\rowY) {\repairedPctInt};
    \node[percentage, text=white]
        at ({0.672+0.00328*(\repairedPctInt+\persistentPct/2)},\rowY)
        {\persistentPct};
}

\node[axis label, xshift=-0.025\linewidth] at (0.464,1) {{Percentage (\%)}};
\node[axis label] at (0.836,1) {{Percentage (\%)}};
\node[panel title, xshift=-0.025\linewidth] at (0.444,-8) {\textbf{(a) Failure Distribution}};
\node[panel title] at (0.836,-8) {\textbf{(b) Missed Violation Distribution}};
\end{scope}

\end{tikzpicture}
\endgroup
\caption{False alarms make up 42--60\% of each agent's failures (a), and for three of the four agents,  most missed violations persist even on scenarios proven to expose the violation (b). A failure is attributed to scenarios only when replacing them with witness scenarios makes the checks reject the violating implementation and accept the gold one; the remaining missed violations are attributed to the checks.}
\label{fig:miss-anatomy}
\end{figure}

A validation harness can be unfaithful in two ways, by raising a false alarm or by missing the violation (Figure~\ref{fig:miss-anatomy}a). These failures involve two components with different obligations. The scenario generator is only required to supply test scenarios that exercise the relevant code, while the checks are scenario-agnostic and must judge correctly on any scenario supplied to them. False alarms therefore arise from the checks. Since a false alarm rejects an implementation the specification permits, it means the checks assert requirements the specification does not impose. These false alarms account for the gap between detection and faithfulness for every agent in Table~\ref{tab:main}. This pattern suggests that agents tend to fit their checks to the reference implementation rather than the specification, a tendency that Section~\ref{sec:ref-ablation} further supports.

Missed violations can arise from insufficient exploration of test scenarios, checks too weak to catch the violation, or both. To separate these causes, we replace the generated scenarios with the differential witness scenarios from Section~\ref{sec:collection} whenever the harness and the witness invoke the same callable in the same form, leave the checks unchanged, and rerun the evaluation. These scenarios distinguish the gold implementation from the violating implementation. A validation harness that still misses the violation therefore fails to detect it on a scenario known to expose it. Among missed violations, the proportion repaired is about two thirds for GPT-5.6-sol and one half for GLM-5.3. For Claude Opus 5 and DeepSeek-V4-Pro, roughly three quarters of missed violations persist with witness scenarios (Figure~\ref{fig:miss-anatomy}b). Even counting every missed violation that the replacement cannot test as a failure of the scenarios, the scenarios account for at most 33--42\% of each agent's failures.

\subsection{Ablations}
\label{sec:ablations}

\subsubsection{Showing the Violating Implementation as the Reference}
\label{sec:ref-ablation}

\begin{table}[t]
\centering
\caption{On the Lite Set, every agent shown the violating implementation as the reference loses 19 to 31 points of faithfulness and rejects the unseen gold implementation on 12--41\% of the tasks. The gold-reference values are the main experiment restricted to these 100 tasks. Gold Rejection is the part of False Alarm on the gold implementation.}
\label{tab:reference}
\small
\begin{tabular}{lcccc}
\toprule
 & \multicolumn{4}{c}{Gold reference\refto violating reference (\%)} \\
\cmidrule(lr){2-5}
Agent & Detection $\uparrow$ & False Alarm $\downarrow$ & Gold Rejection $\downarrow$ & Faithfulness $\uparrow$ \\
\midrule
Codex / GPT-5.6-sol         & 83\refto 78 & 21\refto 37 & \zp4\refto 29 & 65\refto 46 \\
Claude Code / Claude Opus 5 & 66\refto 33 & 19\refto 38 & \zp4\refto 31 & 52\refto 21 \\
OpenCode / DeepSeek-V4-Pro  & 70\refto 67 & 23\refto 48 & 15\refto 41 & 52\refto 28 \\
OpenCode / GLM-5.3          & 75\refto 44 & 12\refto 24 & \zp0\refto 12 & 68\refto 39 \\
\bottomrule
\end{tabular}
\end{table}

In practice, the implementation a validation harness is written against may satisfy its specification or violate it, and a validation harness that merely mirrors whichever implementation it was shown, regardless of what the specification requires, is worthless. To test whether the agents' checks are grounded in the specification rather than the implementation they are shown, we rerun the agents on the Lite Set (Section~\ref{sec:composition}) with the violating implementation as the reference. The first value of each cell in Table~\ref{tab:reference} is the main experiment restricted to the Lite Set. We found that every agent's validation harnesses show a clear tendency to mirror the implementation the agent is given, and that showing the violating implementation as the reference does not help the agent detect it. Faithfulness falls by 19 to 31 points and detection by 3 to 33 points (Table~\ref{tab:reference}). False alarms rise from 12--23\% to 24--48\% of tasks, and most of the increase comes from rejections of the gold implementation, which rise from 0--15\% to 12--41\%.

We further investigate how each agent responds when its own validation harness rejects the reference. Under the violating reference, such a rejection indicates that the implementation violates the specification or that the checks are wrong, and each agent consistently reads nearly all of its own rejections. Claude Opus 5 and GLM-5.3 read them as faults of the checks. As a result, their validation harnesses often accept the implementation they were shown, and their detection rate drops sharply. Moreover, although instructed to follow the specification rather than the implementation, Claude Opus 5 in particular constructs a validation harness that accepts the violating implementation and rejects the gold implementation, showing a clear tendency to follow the reference over the specification. GPT-5.6-sol and DeepSeek-V4-Pro read the rejections as faults of the implementation. Their detection is almost unchanged, but the checks they construct include requirements the specification does not impose, and they now reject the gold implementation, confirming the over-assertion discussed in Section~\ref{sec:failure}.

\subsubsection{Step Budget}
\label{sec:step-budget}

\begin{figure}[t]
\centering
\setlength{\abovecaptionskip}{2pt}
\resizebox{\textwidth}{!}{%
\input{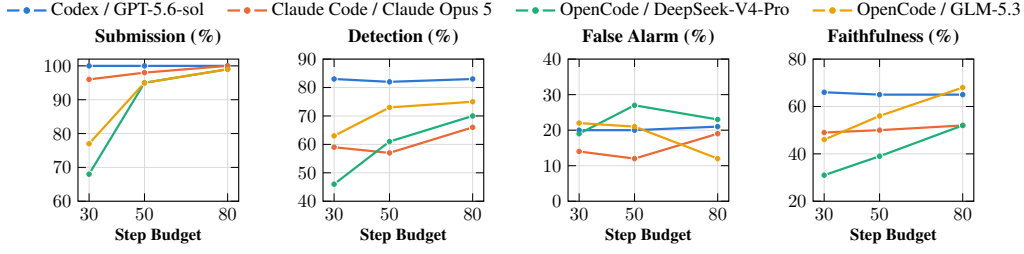}%
}
\caption{Submission rate, detection rate, false alarm rate, and faithfulness rate on the Lite Set under the gold reference, at step budgets of 30, 50, and 80 steps.}
\label{fig:step-budget}
\end{figure}

To test whether an unfaithful validation harness reflects an inability to write the right checks rather than a step budget too small to finish, we compare the 80-step results of the 291-task main experiment, restricted to the Lite Set, with new runs on the Lite Set within 30 and 50 steps. Figure~\ref{fig:step-budget} reports the effect of the budget on each agent. Between 30 and 50 steps, the two open-weight coding agents submit a validation harness on 95\% of the tasks, up from 68\% and 77\%. However, a validation harness submitted within 50 steps is no more likely to detect the violation or to be faithful than one submitted within 30 steps. Among submitted harnesses, both agents become more faithful only between 50 and 80 steps, and for GLM-5.3 the gain comes from fewer false alarms rather than more detections. GPT-5.6-sol is insensitive to the budget at every rate, and the faithfulness of Claude Opus 5 barely changes at any budget.

\section{Conclusion and Discussion}
We have introduced \benchname{}, a benchmark of 291 validation tasks drawn from 30 features of 13 real-world Python repositories, each providing an atomic natural language specification, a reference implementation, a violating implementation that the original test suite cannot catch, and valid alternative implementations that satisfy the specification while differing from the reference. 
The benchmark measures whether a coding agent can construct a validation harness whose checks are faithful to the specification and whose scenarios thoroughly exercise the relevant code, and our evaluation of four proprietary and open-weight coding agents shows that none does so on more than 63\% of the tasks, mainly because their checks assert requirements the specification does not impose and mirror the reference implementation they are shown. 
We hope that \benchname{} will serve as a resource for measuring and improving the ability of coding agents to produce evidence of specification conformance, a capability on which benchmark evaluation, interactive coding workflows, and reinforcement learning with verifiable rewards all depend.

\paragraph{Limitations and future work.}
Despite advancing the evaluation of code validation against natural language specifications, \benchname{} has several limitations.
First, its 291 tasks come from a subset of FeatureBench instances and are unevenly distributed across repositories; future work can extend the automated pipeline to more instances and other programming languages, given various coding benchmarks that provide high-quality specifications and code aligned with them. 
Second, faithfulness is estimated by execution against a finite set of implementations that satisfy or violate the specification, so a check that over-asserts in a way none of them exercises passes undetected; a larger, more diverse pool of alternatives would provide a more precise estimation of the faithfulness.
Finally, the repositories and gold patches are public and may appear in training data, but the violating and alternative implementations are newly synthesized, which mitigates the risk of contamination.

\section*{AI Use Statement}
\label{sec:ai_use}

In this work, we used generative AI tools to assist with language polishing and code debugging. These tools did not contribute to formulating the research ideas or the paper's conclusions. The authors reviewed all AI-assisted content and are responsible for all statements, interpretations, results, and artifacts in this work.

\section*{Reproducibility Statement}
\label{sec:reproducibility}

To ensure reproducibility, our complete code, prompts, dataset, and detailed evaluation settings are available at \href{https://github.com/coxhin/SpecGuardBench}{github.com/coxhin/SpecGuardBench} and \href{https://huggingface.co/datasets/coxhin/SpecGuardBench}{huggingface.co/datasets/coxhin/SpecGuardBench}.

\bibliography{iclr2027_conference}
\bibliographystyle{iclr2027_conference}

\appendix

\section{Verification Logic}
\label{app:logic}

This appendix details the classification of the specifications by verification logic summarized in Section~\ref{sec:composition}. Appendix~\ref{app:logic-procedure} describes how the types were derived and applied, Table~\ref{tab:logic} defines the eight types and gives the counterpart of each in established verification logics, and Appendix~\ref{app:logic-examples} gives one specification of each type.

\subsection{Procedure}
\label{app:logic-procedure}

\paragraph{Open coding.}
Two authors independently coded a sample of 60 specifications, stratified by instance and by specification category, without a predefined set of types. For each specification, they read the specification with its differential witness and recorded in free text what the witness observes to separate the gold implementation from the violating one, and which construct of an established verification logic, if any, expresses the corresponding check. 
The two authors then merged their codes into types by the logical form of the check, which gave the eight types of Table~\ref{tab:logic}.

\paragraph{Closed coding.}
We used Claude Opus 5 to label each of the 291 specifications with exactly one type or as other. Its prompt contains the definitions and boundary rules above, the specification, and its differential witness. To check these labels, the two authors independently labeled a separate stratified sample of 50 specifications with the same definitions and resolved their disagreements by discussion. They agreed on 88\% of the sample before adjudication, and the label of the LLM matched the adjudicated label on 98\% (Cohen's $\kappa$ = 0.97). Two of the 291 specifications were labeled as other.

\begin{table}[h]
\centering
\small
\setlength{\tabcolsep}{4pt}
\caption{The eight types of verification logic, with the counterpart of each type in established verification logics and the number and percentage of the 291 specifications of \benchname{} that belong to each type.}
\label{tab:logic}
\begin{tabular}{@{}>{\centering\arraybackslash}m{3.2cm}>{\raggedright\arraybackslash}m{4.5cm}>{\raggedright\arraybackslash}m{3.8cm}r@{\ }r@{}}
\toprule
Type & A check must establish that & Counterpart & \multicolumn{2}{c@{}}{$n$ (\%)} \\
\midrule
Functional postcondition & the return value satisfies a predicate over the inputs & postconditions (\texttt{ensures} in JML, ACSL, and Dafny) & 143 & (49.1\%) \\
\addlinespace
Input validation & for inputs meeting a stated condition, the call raises a specified exception, returns a specified error value, emits a specified message or warning, or emits none of these & specification cases (JML \texttt{normal\_behavior} and \texttt{exceptional\_behavior}, ACSL behaviors, SPARK \texttt{Contract\_Cases}) & 52 & (17.9\%) \\
\addlinespace
State postcondition & state that outlives the call, in the receiver, an argument, or the module, ends in a required relation to its state before the call or to the inputs & two-state postconditions (\texttt{\textbackslash old} in JML and ACSL, \texttt{old} in Dafny) & 18 & (6.2\%) \\
\addlinespace
Frame condition & the call leaves specified state unmodified, and a result required to be new shares no state with existing objects & frame clauses (\texttt{assignable}, \texttt{assigns}, \texttt{modifies}), \texttt{\textbackslash fresh}, separation logic & 14 & (4.8\%) \\
\addlinespace
Interaction protocol & the function calls specified collaborators with specified arguments and multiplicity, or reads a value from a specified source & I/O specifications in separation logic, flow contracts (SPARK \texttt{Depends}) & 32 & (11.0\%) \\
\addlinespace
Call order & events within one execution occur in a specified order & temporal logic (LTL), typestate & 2 & (0.7\%) \\
\addlinespace
Interface signature & the declared interface has specified parameter names, annotations, default values, or members & type signatures and interface declarations & 23 & (7.9\%) \\
\addlinespace
Relational property & two or more executions stand in a specified relation & $k$-safety hyperproperties, relational Hoare logic & 5 & (1.7\%) \\
\bottomrule
\end{tabular}
\end{table}

\subsection{Examples}
\label{app:logic-examples}

Each example below is a specification of \benchname{} from the named type. For each, we give the check its differential witness performs, written in the notation of the corresponding verification logic, and how the violating implementation of the task departs from it.

\paragraph{Functional postcondition.}
In seaborn, the \texttt{Histogram} statistic states that ``the discrete parameter, when True, automatically sets bin edges to cover integer values with 0.5 offsets.'' For data whose values span the integers $a$ to $b$, a check must establish that the returned edges are $e_i = a - 0.5 + i$ for $i = 0, \ldots, b - a + 1$. The violating implementation keeps the outermost edges and shifts the others by 0.1, which a check of the range or of the number of bins does not detect.

\paragraph{Input validation.}
In astropy, \texttt{default\_cosmology.validate} raises \texttt{TypeError} ``if value is not None, a string, or a Cosmology instance.'' A check must establish that every value outside these three kinds raises \texttt{TypeError}, which in JML is an \texttt{exceptional\_behavior} case whose precondition is the negated type test. The violating implementation also accepts numbers and returns them unchanged. Appendix~\ref{app:example} gives this task in full.

\paragraph{State postcondition.}
In astropy, \texttt{Table.add\_column} states that ``if index is supplied then insert column before index position in the list of columns.'' With primed names for the state after the call, a check must establish $\mathit{colnames}' = \mathit{colnames}[{:}i] + [\mathit{name}] + \mathit{colnames}[i{:}]$, a relation between the order of the columns before and after the call. The violating implementation inserts the column after position $i$ instead.

\paragraph{Frame condition.}
In seaborn, \texttt{Plot.on} states that ``this method returns a copy of the Plot instance, so the original Plot object remains unchanged.'' A check must establish that the call leaves the receiver unchanged, in JML \texttt{assignable \textbackslash nothing} together with \texttt{ensures \textbackslash fresh(\textbackslash result)}, and not only that the result is a new object. The violating implementation returns a copy but also sets the target on the original, so a check that compares only the identity of the result accepts it.

\paragraph{Interaction protocol.}
In MLflow, \texttt{make\_genai\_metric} documents its \texttt{parameters} argument as ``parameters for the LLM used to compute the metric'' and states that ``by default, we set the temperature to 0.0, max\_tokens to 200, and top\_p to 1.0.'' A check must establish that the call to the judge model receives the parameters of the caller when they are given and these defaults otherwise, which is visible only in the arguments of that call and not in the returned metric. The violating implementation overrides the values of the caller with the defaults, so a check that omits \texttt{parameters} accepts it.

\paragraph{Call order.}
In scikit-learn, \texttt{PredictionErrorDisplay} states that ``both from\_estimator and from\_predictions must call check\_matplotlib\_support with the appropriate caller name before performing any operations.'' Over the events of one execution, a check must establish $\neg \mathit{op} \;\mathcal{U}\; \mathit{check}$, where $\mathit{check}$ is the call to \texttt{check\_matplotlib\_support} and $\mathit{op}$ is any other operation. The violating implementation makes the same calls in the reverse order and returns the same result.

\paragraph{Interface signature.}
In seaborn, \texttt{Plot.label} documents its parameter \texttt{title} as ``str or None, optional.'' A check must establish that the declared annotation of \texttt{title} admits \texttt{None}, which a type checker reads from the signature without running the method. The violating implementation narrows the annotation to \texttt{str} and behaves identically.

\paragraph{Relational property.}
In mypy, the type-inference class \texttt{Constraint} requires hash values that are ``consistent with equality semantics for use in collections and constraint solving.'' A check must establish $c_1 = c_2 \Rightarrow \mathit{hash}(c_1) = \mathit{hash}(c_2)$, a relation between two executions that no single execution decides. The violating implementation hashes a name that equality ignores, so two equal constraints built from distinct type-variable objects hash differently.

\section{Feature-Level Results}
\label{app:feature-level}

The results of Tables~\ref{tab:main} and~\ref{tab:reference} also hold when every feature is given equal weight. Each FeatureBench instance targets one feature and contributes a different number of atomic specifications, from 2 to 41 in the Full Set and from 1 to 14 in the Lite Set. The two tables average over tasks and therefore weight each feature by its number of tasks. Tables~\ref{tab:main-feature} and~\ref{tab:reference-feature} report the same metrics averaged over features instead, computing each metric within each of the 30 features and then taking the unweighted mean over the features.

\begin{table}[h]
\centering
\caption{Performance of the four coding agents on the 291 validation tasks of \benchname{}, averaged over the 30 features instead of over tasks. Each metric is defined as in Section~\ref{sec:task}, computed within each feature, and then averaged over the features with equal weight. Cost denotes the average cost per task in U.S. dollars.}
\label{tab:main-feature}
\small
\begin{tabular}{lccccc}
\toprule
Agent & Detection $\uparrow$ & False Alarm $\downarrow$ & Faithfulness $\uparrow$ & Edge Cov. $\uparrow$ & Cost \\
\midrule
Codex / GPT-5.6-sol         & \textbf{83.0\%} & 24.3\% & \textbf{63.4\%} & 82.4\% & \$0.73 \\
Claude Code / Claude Opus 5 & 62.6\% & 19.9\% & 50.5\% & 73.9\% & \$1.76 \\
OpenCode / DeepSeek-V4-Pro  & 64.7\% & 25.5\% & 45.1\% & 73.7\% & \$0.43 \\
OpenCode / GLM-5.3          & 75.5\% & \textbf{17.2\%} & 62.5\% & \textbf{84.4\%} & \$1.41 \\
\bottomrule
\end{tabular}
\end{table}

Averaged over features, every metric of the main experiment changes by less than three points (Table~\ref{tab:main-feature}). The agents keep their order on detection and on false alarms, faithfulness still ranges from 45\% to 63\%, and detection exceeds faithfulness by 12 to 20 points for every agent. GPT-5.6-sol and GLM-5.3 are again one point apart on faithfulness, now in the opposite order, and remain the two agents with the highest edge coverage.

\begin{table}[h]
\centering
\caption{Table~\ref{tab:reference} averaged over the 30 features of the Lite Set instead of over its 100 tasks. Each metric is computed within each feature and then averaged over the features with equal weight.}
\label{tab:reference-feature}
\small
\begin{tabular}{lcccc}
\toprule
 & \multicolumn{4}{c}{Gold reference\refto violating reference (\%)} \\
\cmidrule(lr){2-5}
Agent & Detection $\uparrow$ & False Alarm $\downarrow$ & Gold Rejection $\downarrow$ & Faithfulness $\uparrow$ \\
\midrule
Codex / GPT-5.6-sol         & 86.5\refto 78.3 & 24.8\refto 31.7 & \zp1.5\refto 24.1 & 66.2\refto 51.5 \\
Claude Code / Claude Opus 5 & 64.3\refto 40.6 & 19.8\refto 35.8 & \zp5.3\refto 28.7 & 49.6\refto 25.2 \\
OpenCode / DeepSeek-V4-Pro  & 75.7\refto 72.2 & 22.2\refto 51.6 & 16.4\refto 43.1 & 57.3\refto 30.6 \\
OpenCode / GLM-5.3          & 80.5\refto 49.2 & 14.2\refto 29.2 & \zp0.0\refto 13.3 & 70.9\refto 42.1 \\
\bottomrule
\end{tabular}
\end{table}

Showing the violating implementation as the reference has the same effect when features are weighted equally (Table~\ref{tab:reference-feature}). Every agent loses 15 to 29 points of faithfulness, and rejections of the gold implementation rise from 0--16\% to 13--43\%. Detection falls by 24 and 31 points for Claude Opus 5 and GLM-5.3 but only by 8 and 4 points for GPT-5.6-sol and DeepSeek-V4-Pro, the same split between the two pairs of agents as in Section~\ref{sec:ref-ablation}. These averages are coarser than those over the Full Set, because 8 of the 30 features contribute a single task to the Lite Set and enter every average at either 0\% or 100\%.

\begin{table}[h]
\centering
\caption{Table~\ref{tab:main} with 95\% confidence intervals clustered by feature. The last three rows give the differences between the agents that Section~\ref{sec:main-results} compares.}
\label{tab:main-ci}
\small
\begin{tabular}{lcccc}
\toprule
Agent & Detection & False Alarm & Faithfulness & Edge Cov. \\
\midrule
Codex / GPT-5.6-sol               & 81.8 $\pm$ 5.9 & 23.0 $\pm$ 5.7 & 61.9 $\pm$ 8.0 & 82.0 $\pm$ 6.4 \\
Claude Code / Claude Opus 5       & 65.3 $\pm$ 5.4 & 20.6 $\pm$ 4.5 & 50.2 $\pm$ 6.3 & 72.4 $\pm$ 6.6 \\
OpenCode / DeepSeek-V4-Pro        & 66.7 $\pm$ 5.8 & 26.8 $\pm$ 7.1 & 45.0 $\pm$ 6.1 & 75.1 $\pm$ 7.7 \\
OpenCode / GLM-5.3                & 74.2 $\pm$ 5.8 & 15.1 $\pm$ 4.6 & 62.9 $\pm$ 6.9 & 84.3 $\pm$ 4.7 \\
\midrule
GPT-5.6-sol $-$ GLM-5.3           & $+$7.6 $\pm$ 6.6 & $+$7.9 $\pm$ 4.4 & $-$1.0 $\pm$ 6.8 & $-$2.3 $\pm$ 4.1 \\
DeepSeek-V4-Pro $-$ Claude Opus 5 & $+$1.4 $\pm$ 5.4 & $+$6.2 $\pm$ 5.4 & $-$5.2 $\pm$ 5.5 & $+$2.7 $\pm$ 4.2 \\
Two most faithful $-$ other two   & $+$12.0 $\pm$ 4.7 & $-$4.6 $\pm$ 5.0 & $+$14.8 $\pm$ 4.9 & $+$9.4 $\pm$ 3.3 \\
\bottomrule
\end{tabular}
\end{table}

\section{Terms and Notation}
\label{app:terms}

The contracts in Appendix~\ref{app:task} are quoted verbatim. They keep the names the pipeline uses internally, which differ from the terms of the paper; Table~\ref{tab:terms} maps one to the other. Prompts are Jinja templates. A field written as \texttt{\{\{ x \}\}} is filled per call, and a block delimited by \texttt{\{\% for \%\}} or \texttt{\{\% if \%\}} is expanded per call. Both are shown as they appear in the template. Em dashes and arrows in the original files are typeset as \texttt{--} and \texttt{->}.

\begin{table}[h]
\centering
\small
\caption{Names used inside the pipeline and the corresponding terms of the paper.}
\label{tab:terms}
\begin{tabular}{ll}
\toprule
In the contracts & In the paper \\
\midrule
certificate & validation harness \\
\texttt{build\_cases()}, coverage cases & scenario generator, test scenarios \\
\texttt{check\_case(case)} & executable checks \\
case & test scenario \\
\texttt{target} & the callable the scenarios invoke \\
gold implementation, gold patch & gold implementation \\
mutant, mutation & violating implementation \\
control mutant & valid alternative implementation \\
other specs, sibling specs & the other atomic specifications of the instance \\
LLM-A & the synthesizing LLM (Claude Opus 5) \\
witness factory, observer & differential witness \\
preservation probe & a neighbor's witness run against the candidate \\
fired, silent & rejects, accepts \\
\bottomrule
\end{tabular}
\end{table}

\section{Task Interface}
\label{app:task}

This appendix gives what the agent receives in a task. Appendix~\ref{app:harness} is the guide that fixes the form of a validation harness, Appendix~\ref{app:case-contract} the contract every test scenario must satisfy, and Appendix~\ref{app:example} one complete task of \benchname{} together with a validation harness submitted for it.

\subsection{Harness Contract}
\label{app:harness}

The agent is instructed to read the following guide from the repository root before working. It fixes the two callables of a validation harness, the shape of a test scenario, and how the callable under validation is declared. In the terms of the paper, \texttt{build\_cases()} is the scenario generator and \texttt{check\_case(case)} the executable checks.

\begin{lstlisting}
# Factory-Oracle Harness Guide

This harness separates input construction from the frozen specification oracle. Your certificate is a Python module with exactly two public callables:

```python
def build_cases():
    """Yield baseline-generated cases used to measure edge coverage."""
    yield {"args": (...,), "kwargs": {...}}


def check_case(case):
    """Assert the requested spec for one case."""
    ...
```

The evaluator freezes this module after generation. It calls `build_cases()` and runs `check_case(case)` on the provided implementation to measure edge coverage. The same frozen certificate may also be used to test other implementations and may receive additional cases that satisfy the case contract.

## Hard constraints

- Do not edit source files under `/testbed`.
- `build_cases()` must return fresh Python cases and must not contain assertions.
- Every case must be exactly a dict with `args` and `kwargs`. `args` is a tuple or list, and `kwargs` is a dict with string keys. Invoke the target as `target(*case["args"], **case["kwargs"])`.
- Expose the callable your cases invoke as a module-level global named `target`, bound directly to an imported function or method rather than a certificate-defined wrapper. You choose which callable that is. If `target` is an instance method taken from a class, include the receiver instance as the first positional argument; a method taken from an instance is already bound and does not need that extra argument.
- `target` is the callable you declared, unmodified. Evaluation does not substitute, wrap, or instrument it, so anything true of it in your own environment is true during evaluation.
- Declaring `target` constrains only what a case *is* -- how you observe is your choice: `check_case` may additionally import internals, wrap them with `unittest.mock.patch(..., wraps=...)`, or inspect side effects to judge the specification.
- `check_case(case)` must work for every case satisfying the supplied case contract, not only the cases returned by your own `build_cases()`.
- A violation can manifest in many ways, for example through a return value, an exception, a side effect, and so on. Determine from the specification itself which aspects of the target's observable behavior it constrains. Judge the actual behavior against the specification for the case; do not assume the target always returns normally.
- Handle target exception behavior according to the specification. If an exception makes the target violate the specification for the case, express that violation as an assertion failure. Keep the `try` block narrowly scoped to target invocation so certificate-oracle bugs are not hidden.
- Cases describe stimulus only. Do not store expected outcomes in case keys such as `expected`, `expected_result`, `should_raise`, or `expected_calls`.
- Keep all expected-behaviour logic inside `check_case`.
- Do not compare against a copied reference implementation.
- Do not inspect source text, bytecode, mutation patches, or evaluation state.
- Do not use Hypothesis or time-bounded generation in the final certificate.
- You may use monkeypatching when it helps exercise additional edges.
- Factories may construct real Python objects, dynamic classes, temporary resources, callbacks, or multi-step scenario objects when required by the contract.
- For setup/teardown, implement `build_cases()` as a generator with `try/finally`. The harness closes the generator after execution, including when `check_case` fails.

## Example

```python
from package.module import function_under_test


target = function_under_test


def build_cases():
    yield {"args": (0,), "kwargs": {}}
    yield {"args": (1,), "kwargs": {}}


def check_case(case):
    result = target(*case["args"], **case["kwargs"])
    assert result >= 0
```
\end{lstlisting}

\subsection{Case Contract}
\label{app:case-contract}

Every task carries the same case contract that constrains the form of a test scenario, not the values it may hold. Argument values are bounded only by the specification and the public calling interface of the callable under validation, so the checks must judge any scenario within those bounds, including scenarios the agent did not generate.

\begin{lstlisting}
Every case is a dict with exactly two keys:

- `args`: a tuple or list of positional arguments for the target callable.
- `kwargs`: a dict mapping string keys to keyword argument values for the target callable.

`check_case(case)` must invoke the target using `target(*case["args"], **case["kwargs"])`. The evaluator may provide unseen cases using the same representation. Argument values are constrained only by the natural-language specification and the target's public calling interface. The target must be the actual implementation callable, not a wrapper invented by the witness or certificate. If an instance method is taken from its class, `args` includes the receiver instance first; a method already taken from an instance is bound and does not need that extra argument.
Cases may exercise normal, boundary, or error-handling behavior when that behavior is covered by the specification or public interface. Whether an exception or any other observed behavior is a violation depends on what the specification requires for that case.
Cases contain stimulus only and must not contain expected outcomes, expected exceptions, failure flags, or mutant-specific information.
\end{lstlisting}

\subsection{A Complete Task}
\label{app:example}

The task below comes from an astropy instance whose feature includes \texttt{default\_cosmology}, the setting that selects the default cosmology. The atomic specification is one sentence of the docstring of \texttt{default\_cosmology.validate}. The violating implementation widens the type check so that numeric values pass through instead of raising, and the original test suite, which reaches this branch only with a class object and with cosmology instances, does not catch it.

\begin{lstlisting}
Instance: astropy__astropy.b0db0daa.test_realizations.5b9cf523.lv1
Specification S16 (behavioral): Raises TypeError: If value is not None, a string, or a Cosmology instance.
Implemented at: astropy/cosmology/_src/default.py, lines 66-71, default_cosmology.validate
\end{lstlisting}

\paragraph{Violating implementation.}\mbox{}
\begin{lstlisting}
diff --git a/astropy/cosmology/_src/default.py b/astropy/cosmology/_src/default.py
index 48ca309529..b4f55d277d 100644
--- a/astropy/cosmology/_src/default.py
+++ b/astropy/cosmology/_src/default.py
@@ -66,7 +66,7 @@ class default_cosmology(ScienceState):
                 value = None
             else:
                 value = cls._get_from_registry(value)
-        elif not isinstance(value, Cosmology):
+        elif not isinstance(value, (Cosmology, int, float)):
             raise TypeError(
                 "default_cosmology must be a string or Cosmology instance, "
                 f"not {value}."
\end{lstlisting}

\paragraph{Differential witness.}
The witness consists of a factory that yields scenarios and an observer that invokes the callable on one scenario and returns what it observed. The pipeline runs the observer on every scenario of the factory against the gold implementation and against the candidate and compares the returned observations. Here the gold implementation raises \texttt{TypeError} on every scenario and the candidate returns the value unchanged.

\begin{lstlisting}[language=Python]
def build_cases():
    """Yield calls to default_cosmology.validate with values that are neither
    None, a string, nor a Cosmology instance."""
    for value in (1, 0, -3, 2.5, True):
        yield {"args": (value,), "kwargs": {}}
\end{lstlisting}
\begin{lstlisting}[language=Python]
from astropy.cosmology.realizations import default_cosmology

target = default_cosmology.validate


def observe_case(case):
    args, kwargs = tuple(case["args"]), dict(case["kwargs"])
    try:
        result = target(*args, **kwargs)
    except Exception as exc:  # noqa: BLE001 - the exception type is the observation
        return {
            "raised": type(exc).__name__,
            "mentions_string_or_cosmology": "must be a string or Cosmology" in str(exc),
        }
    return {
        "raised": None,
        "result_type": type(result).__name__,
        "result_repr": repr(result),
    }
\end{lstlisting}
\begin{lstlisting}
# gold implementation
{"mentions_string_or_cosmology": true, "raised": "TypeError"}
{"mentions_string_or_cosmology": true, "raised": "TypeError"}
{"mentions_string_or_cosmology": true, "raised": "TypeError"}
{"mentions_string_or_cosmology": true, "raised": "TypeError"}
{"mentions_string_or_cosmology": true, "raised": "TypeError"}

# violating implementation
{"raised": null, "result_repr": "1", "result_type": "int"}
{"raised": null, "result_repr": "0", "result_type": "int"}
{"raised": null, "result_repr": "-3", "result_type": "int"}
{"raised": null, "result_repr": "2.5", "result_type": "float"}
{"raised": null, "result_repr": "True", "result_type": "bool"}
\end{lstlisting}

\paragraph{A submitted validation harness.}
The following harness was submitted by Codex with GPT-5.6-sol in the main experiment. It rejects the violating implementation and accepts the gold implementation and its valid alternatives, and is therefore counted as faithful. The scenario generator mixes values that must raise a \texttt{TypeError} with \texttt{None}, strings, and a Cosmology instance, on which the specification imposes nothing. The checks decide from the scenario alone whether the specification requires a \texttt{TypeError} and assert nothing otherwise, so they apply to scenarios other than the generated ones.

\begin{lstlisting}[language=Python]
import inspect

from astropy.cosmology import Cosmology
from astropy.cosmology.realizations import Planck18, default_cosmology


target = default_cosmology.validate


def build_cases():
    invalid_values = [
        0,
        1.0,
        False,
        b"Planck18",
        [],
        {},
        object(),
        TypeError,
    ]
    for value in invalid_values:
        yield {"args": (value,), "kwargs": {}}

    yield {"args": (), "kwargs": {"value": None}}
    yield {"args": ("no_default",), "kwargs": {}}
    yield {"args": ("Planck18",), "kwargs": {}}
    yield {"args": ("not-a-real-cosmology",), "kwargs": {}}
    yield {"args": ("available",), "kwargs": {}}
    yield {"args": (Planck18,), "kwargs": {}}


def check_case(case):
    arguments = inspect.signature(target).bind(*case["args"], **case["kwargs"])
    value = arguments.arguments["value"]
    requires_type_error = (
        value is not None
        and not isinstance(value, str)
        and not isinstance(value, Cosmology)
    )

    if requires_type_error:
        try:
            target(*case["args"], **case["kwargs"])
        except TypeError:
            return
        except BaseException as error:
            raise AssertionError(
                f"invalid value must raise TypeError, not {type(error).__name__}"
            ) from error
        raise AssertionError("invalid value did not raise TypeError")

    try:
        target(*case["args"], **case["kwargs"])
    except BaseException:
        pass
\end{lstlisting}

\end{document}